\documentclass[10pt,twocolumn,letterpaper]{article}

\usepackage{iccv}
\usepackage{times}
\usepackage{epsfig}
\usepackage{graphicx}
\usepackage{amsmath}
\usepackage{amssymb}

\usepackage[pagebackref=true,breaklinks=true,letterpaper=true,colorlinks,bookmarks=false]{hyperref}

\iccvfinalcopy 

\ificcvfinal

\begin{document}


\pagenumbering{arabic}
\pagestyle{plain}

\title{Follow the Attention: Combining Partial Pose and Object Motion for Fine-Grained Action Detection}

\author{M. Mahdi Kazemi M.\\
\and
Ehsan Abbasnejad\\
\and
Javen Shi\\
The Australian Institute for Machine Learning, The University of Adelaide\\
{\tt\small mohammadmahdi.kazemimoghaddam, ehsan.abbasnejad, javen.shi@adelaide.edu.au }
}

\maketitle
\begin{abstract}
   Retailers have long been searching for ways to effectively understand their customers' behaviour in order to provide a smooth and pleasant shopping experience that attracts more customers everyday and maximises their revenue, consequently. Humans can flawlessly understand others' behaviour by combining different visual cues from activity to gestures and facial expressions. Empowering the computer vision systems to do so, however, is still an open problem due to its intrinsic challenges as well as extrinsic enforced difficulties like lack of publicly available data and unique environment conditions (wild).
   In this work, We emphasise on detecting the first and by far the most crucial cue in behaviour analysis; that is human activity detection in computer vision. To do so, we introduce a framework for integrating human pose and object motion to both temporally detect and classify the activities in a fine-grained manner (very short and similar activities). We incorporate partial human pose and interaction with the objects in a multi-stream neural network architecture to guide the spatiotemporal \emph{attention} mechanism for more efficient activity recognition. To this end, in the absence of pose supervision, we propose to use the Generative Adversarial Network (GAN) to generate exact joint locations from noisy probability heat maps. Additionally, based on the intuition that complex actions demand more than one source of information to be identified even by humans, we integrate the second stream of object motion to our network as a prior knowledge that we quantitatively show improves the recognition results. We empirically show the capability of our approach by achieving state-of-the-art results on MERL shopping dataset. We further investigate the effectiveness of this approach on a new shopping dataset that we have collected to address existing shortcomings.
\end{abstract}

\begin{figure}[t]
\begin{center}
   \includegraphics[width=\linewidth]{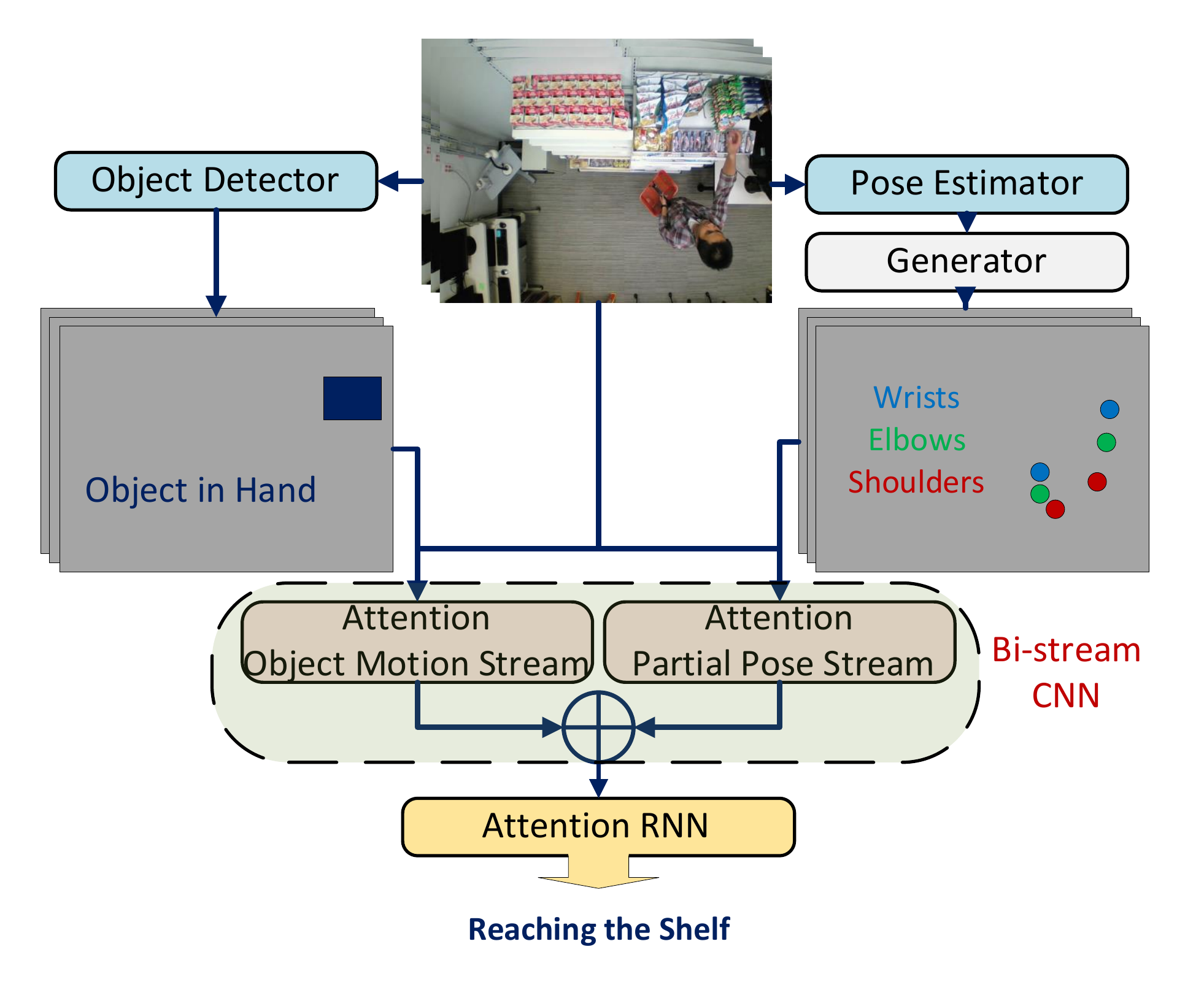}
\end{center}
   \caption{Black box overview of our proposed method.}
\label{fig:front_page}
\end{figure}

\section{Introduction}

Recent advances in artificial intelligence (AI) have had a dramatic impact on various technologies that affect our everyday lives. In areas such as retail and shopping, however, this impact has remained limited. More efficient surveillance, customer behaviour analysis, targeted marketing etc. are among the benefits of employing such technologies in this environment. Deployment of these methods will provide more efficiency and higher revenue to the retailers as well as more pleasant shopping experience to the customers. Furthermore, the drastically growing demand for these technological solutions can be inferred from the share of the global trade market these technologies hold. According to research carried out by Grand View Research organisation, the market value of these AI-based technologies in the retail sector is estimated to reach almost US\$10 billion by the year 2025.

Applying AI, and machine learning methods in particular, in the shopping area is still significantly challenging partly because of the lack of data mainly due to the privacy concerns, costly labelling and remaining proprietary where obtained. Even when datasets are publicly available to the research community (MERL shopping dataset~\cite{merl}) applying deep learning methods on them is challenging due to the additional extrinsic challenges proposed by the unique environmental conditions such as video quality, camera view angle, complex interactions between customers and products and high occlusion. Nevertheless, the recent success of deep learning approaches is partly owed to the use of large scales public datasets such as ImageNet~\cite{imagenet}, UCF101~\cite{ucf101} or \cite{pose_ds} which allow complex models with thousands of parameters to be optimally trained. Even in domains with smaller datasets, models learned on large ones can be \emph{transferred} to facilitate training in a new domain. In shopping scenarios where such datasets are unavailable or the environmental conditions, \eg camera viewing angle, are very different from the data that the existing models are trained on, we need more sophisticated and powerful algorithms to compensate for these shortcomings and overcome the challenges.
The main visual cue in understanding human behaviour is the performed activities which then combined with other cues such as facial expression while tracked over periods of time can provide in-depth behaviour understanding~\cite{behav1, behav2, behav3}. Here we focus on the task of \emph{activity recognition and detection} from shopping surveillance video inputs as the fundamental component. In recognition, we are concerned with only classifying the given trimmed video into available classes of activities while during the detection classification is applied to a continuous sequence of videos of multiple activities; that is the starting temporal location and the duration of the activities are also unknown and desired. This is particularly essential in order to be applicable to the shopping environment. In particular, we find the following specific problems that need to be addressed before effectively achieving the desired task on shopping video data: (1) based on the observation that human pose and object (\eg product) interaction underpin the activities of a customer in a shopping environment, we require a special deep learning architecture including spatiotemporal attention mechanism to attend to various regions of the input frames and their association to others in a sequences; (2) even though \emph{transfer learning} as a common practice in current computer vision community is not directly applicable for pose and object detection in shopping environment, using these knowledge priors are still required; and, finally, (3) new datasets with challenging sequences of activities that more realistically reflect the true shopping scenarios are also desirable for optimal training of complex architectures. 

\begin{figure*}
\begin{center}
   \includegraphics[width=\textwidth]{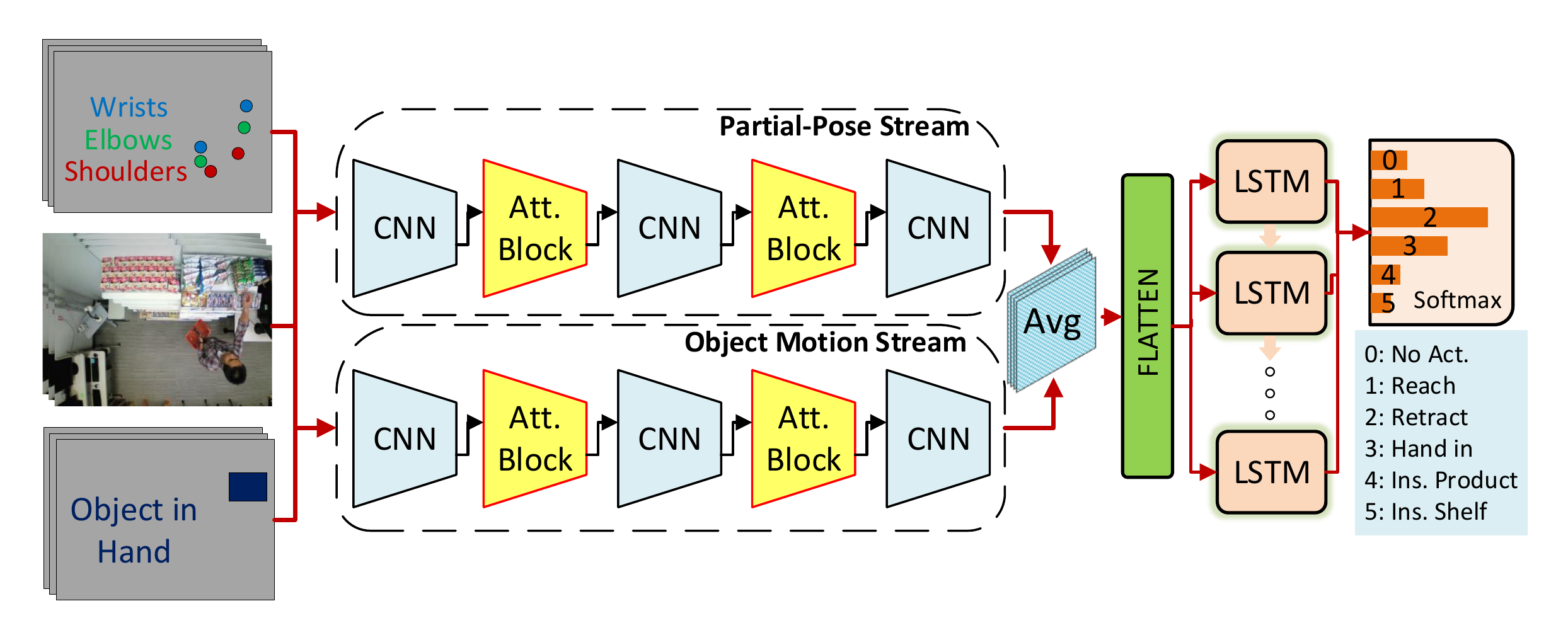}
\end{center}
   \caption{Approach overview: our proposed bi-stream attentive recurrent neural network.}
\label{fig:overview}
\end{figure*}

In Figure~\ref{fig:overview} we summarise our proposed framework to tackle the above-mentioned key issues and many other minor ones. For (1), we propose a bi-stream recurrent neural network architecture to guide the attention using the estimated customers' body pose and object-of-interest motion. Furthermore, inspired by the work introduced in \cite{att}, we devise an attention mechanism to combine human pose and raw input frame information to attend to regions of interest that correspond to a particular activity. These regions are detected in a correlated way both in the spatial and temporal domain. This approach leads to a very effective spatiotemporal attention module that aligns well with the shopping environment. For (2), our experiments show that the state-of-the-art pre-trained pose estimation and object detection methods do not generalise well to our task due to the unique camera view angle in current dataset (top-view) discussed before. Additionally, obtaining critical human pose annotations on the dataset at hand, though looking straightforward, is a tiresome resource-hungry task. Therefore, we devise a generative adversarial network (GAN) to produce potential body joint locations in an unsupervised manner for the unconventional camera view angle.

In summary, Our contributions can be listed as follows:
\begin{itemize}
\item We propose a novel method for utilising partial body pose in the absence of accurate joint locations, empowered by GANs, for an under-explored camera view-angle (top view).
\item We propose a simple, yet effective, algorithm in order to use our action recogniser network as an action detector that temporally detects action locations while classifying them. This method utilises two stream network for combining various sources in semantic space and conquers the complicated detection task by dividing into recognition and detection modules. 
\item We effectively integrate two different sources of information for action 
    recognition/ detection; namely human body pose (partial) and object-of-interest motion to encourage more attention and network resources to essential available information while discarding the less important signals. This is accomplished by utilising \emph{self-attention} where the important spatial regions of the video are correlated to other regions in neighbouring frames for improved final results in terms of accuracy and/ or precision.
\item We achieve state-of-the-art results on MERL shopping dataset with its 
    unique challenges like camera view angle, activity classes and small training data utilising our proposed novel method in conjunction with the traditional transfer learning method.
\item And finally, we release a new dataset dedicated to shopping activities including all its challenges to encourage future research in this area by both increasing the amount of available training data as well as addressing some existing drawbacks that we spotted during experiments. This dataset will be publicly accessible.
\end{itemize}

The rest of the thesis is organised as follows: in section~\ref{our_approach} we discuss each major component of our approach in detail; in section~\ref{exp_ablation} we provide extensive experimental setup details as well as results and ablation study on the two datasets (\eg MERL and AIML); this empirically proves the effectiveness and integrity of the proposed approach. Next, in section~\ref{RL_att} we build grounds for a novel extension idea that is in the implementation phase; despite being a work-in-progress we provide arguments defending the novelty, expected contribution and the achievability of the proposed method. We end this thesis by providing a list of ideas to encourage future work in this area. It should be noted that these ideas are based on insights gained during experiments and seem reasonable to invest on.

\subsection{Related Works}
\subsubsection{Activity Recognition}

\textbf{Hand-Crafted Features}
This body of research is essentially concerned with engineering feature extraction methods that can efficiently detect and represent motion in the input sequence of frames. There have been many methods introduced from early stages of space-time pyramids~\cite{handcr1} till recent years of combining optical flow (OF) and feature matching~\cite{handcr2}; with the most recent ones as attempts to substitute deep neural network feature extractors.

Estimating optical flow from successive video frames, in particular, however, has been proven very effective. A plethora of methods exists in the literature surrounding this topic. One of the earliest methods was introduced by Horn \etal~\cite{horn_of}. As has been the case for many hand-crafted features in recent years, there exist attempts to combine these methods with deep neural networks aiming for the best-of-both-worlds results. Examples of such include~\cite{flownet:_2015, flownet2_2016}, where the estimation of optical flow is observed as a supervised task handled by Convolutional Neural Networks(CNNs). On the other side, there are methods that mostly follow the traditional hand-crafted methods of OF estimation while taking advantage of feature extraction by CNNs. The latter has been proven more effective in some cases~\cite{pwc-net:_2017}. There are also other CNN-based approaches that observe the OF estimation as an unsupervised task to compensate for the lack of annotations as well as issues associated with synthetically generated datasets~\cite{geonet:_2018, unflow:_2017, unflow_occlusion_2017}.

\textbf{Pose-Based Activity Recognition} 
Recent significant progress in pose estimation from single RGB images~\cite{convolutionalpose, pose, stacked_2016, densepose:_2018, simple_2018, articulated-pose} has motivated the utilisation of it as a high-level feature that effectively describes different types of activities~\cite{2d/3d_2018, online_2019} in videos. However, although relatively accurate in terms of single image estimation, the slightest inaccuracy or a single perturbation in sequential pose estimation in videos is detrimental to the task of activity understanding using current state-of-the-art algorithms. As we'll empirically show, this happens quite often even under normal less challenging conditions in terms of missing joint locations in frames. Therefore, many approaches have been proposed to utilise this valuable information as an auxiliary source of information in conjunction with other sources like optical flow, raw input frame etc. One interesting way to address this is to use the by-product of pose estimation; that is the heat map of joint locations used to build the full-body pose in pose-processing~\cite{heat_action}. Although from this perspective our approach appears to be similar, there exist major differences. The authors solely rely on pose heat maps as an individual source of information while we treat them as a guiding source to our network to extract higher level action related features from the human body. Also, we boost the performance of pose estimation using GANs, while in their case heat maps are less noisy and more reliable, thus directly used as inputs.

Furthermore, there exist other approaches to benefit from body pose features while ignoring the inaccuracies. Some define hand-crafted methods to encode the sequential pose information into images for action classification~\cite{joint-traj, approach_posebased}.

If the pose estimation is accurate enough (\eg 3D pose using depth camera or motion capture skeletal sensors), there are enormous ways to effectively incorporate them into activity understanding~\cite{3d-skeletons, normal_vec, space-time-rep, hierarchical-rnn}. However, the use of sensor fusion methods or depth cameras is not feasible in the shopping environment due to low range depth, crowd and occlusion as well as expenses. 
Other methods include the use of graph neural networks by defining the body pose as a graph with joints as nodes and attributes describing the relative speed, angle and location of each node to help activity understanding~\cite{graph-cnn, deep-progressive-rl}.
The other aspect of shopping activities is the number of object interactions that renders it impossible to solely rely on high-level pose information to distinguish them.

\textbf{Object-related actions} Human-object interaction is a long-standing problem in computer vision. The type of considered interactions, however, are mostly around sports activities~\cite{mutual-context}, cooking~\cite{composite-activity, 50salads} or everyday activities~\cite{ucf101, grouplet, hmdb, learn-to-detect}; that could sometimes even be classified from single images~\cite{spatial-functional}. We argue that the type of interactions in the shopping environment is quite different. Firstly, all the objects (\eg various products) could be considered a single category of the object without hurting the recognition. Secondly, only the movement of them in the scene helps distinguish some activities (picking from shelf vs putting back). This is why we name this category of activities as object-related rather than interactions. Kim \etal~\cite{kim} propose an efficient method to recognise actions involving objects. The authors use graph neural networks to fuse human pose and object pose data for action recognition from surveillance cameras. However, their approach is heavily reliant on the quality of input pose information. On the contrary, our proposed approach is more robust to inaccurate input pose.  

\textbf{Multi-stream neural networks}
Recently, the use of multi-stream CNNs has become a prevalent practice to combine different modalities of information before decision making~\cite{segment_multi-stream, two_str_ziss1, ziss2, merl, action-localization, of-guided, ts-lstm}. In most of the introduced approaches, however, it can be observed that these approaches allocate a specific stream for temporal understanding, mostly using a previously trained and calculated optical flow algorithm~\cite{flownet:_2015}, while another dedicated to digging into spatial details of the scene. The novelty of our approach hence is the different modality of streams. That is, firstly, both streams of our network capture the spatial and temporal dependency among input frame sequences. Secondly, the two streams are both augmented with an attention mechanism. Therefore, we combine two modalities of information that can individually represent an activity while acting complementary to improve the accuracy of recognition. 

\textbf{3D-CNNs} The success of 2D CNNs in image understanding has motivated many researchers to exploit the possibility of accomplishing the same in videos. This has led to the introduction of many methods that try to expand the convolution in time~\cite{ziss3d, end3d,3d-cnn,spatiotemporal-3dcnn}. The problem with these methods is that they are still very computationally expensive while only performing slightly better than 2D CNN counterparts. This has led many researchers to rethink the use of them and invent smart combinations of both 2D and 3D CNNs to get the best of both worlds~\cite{rethink_spatiotemp}. For the task of using pose information, however, to the best of our knowledge, the area is still under-explored.
\subsubsection{Activity Detection}
When it comes to real-world applicability, although many powerful methods have been introduced they most make a big assumption. That is, the temporal location of the activity is somehow known. This doesn't hold in many cases and our shopping environment is not an exception. Therefore, there are methods that only address the temporal detection aspect of the task while others combine the two tasks and address them simultaneously~\cite{temporal-detection,single-shot,cdc,action-localization,context-net}. Also known as segmentation, some are inspired by object detection methods and use temporal region proposals~\cite{rethink-faster,turn-tap,rc3d,cascade-proposals}. In our approach, we argue that developing an efficient enough activity recogniser is good enough to precisely detect activities using the simplest of detection methods such as the averaging sliding window we used here.
\subsubsection{Attention Mechanism} Inspired by the psychological meanings of human attention, attention mechanisms are becoming an inseparable part of neural networks when it comes to a task like image or video understanding. That is because concentrating on salient regions in input frames can significantly improve the final results. Initially started in the natural language processing community on machine translation tasks(machine translation), many novel methods have been introduced to mimic human attention in CNNs. Some of them only focus on spatial regions of importance~\cite{study-spatial-att,attention-allneed,transformer,bottom-up}, while others are used in sequential tasks to emphasise on important input frames~\cite{attention-aware-rl,end-to-end-att}. Wang \etal~\cite{att} and integrate both in a single plug-in differentiable module. It has gone beyond this to use complex algorithms like Reinforcement Learning (RL) to define an agent that can find the salient regions or frames~\cite{attention-aware-rl,deep-progressive-rl}. 

\subsubsection{Available Datasets}
Many datasets have been introduced to help develop and test algorithms in this body of research~\cite{ava, hollywood, ucf101, act-net}. Kinetics~\cite{vadis} is one of the largest ones that is considered the ImageNet~\cite{imagenet} of the videos. Among these only a few consists of untrimmed videos that allow detection~\cite{composite-activity,50salads, thumos}. More importantly, as mentioned before, the proprietary nature of retail-related datasets means the lack of data to experiment on algorithms that try to overcome the unique challenges. To the best of our knowledge, MERL~\cite{merl} dataset is the only available dataset on our task at hand with very limiting characteristics such as single person, sparse and unrealistic activities and limited training data. 

\section{Our Approach} \label{our_approach}
\subsection{Unsupervised Fine-Tuning of Pose Network}
Generative Adversarial Networks (GANs) have been proven useful in many semi-supervised and unsupervised tasks recently. The intuitive idea behind this family of neural networks is based on a min-max algorithm, in which two parties compete to maximise their own (contradictory) reward functions. A GAN consists of two models: a generative model which is trained to learn the probability distribution of the input data and a discriminative model which tries to distinguish real samples of the input from fake generated ones. The two models can be trained simultaneously using back-propagation.

The success of GANs can easily be inferred from the everyday increasing application of them ranging from artistic image generation and super-resolution to semi-supervised classification. However, to the best of our knowledge, using them as a substitute for troublesome and resource-consuming (especially in the absence of annotations) task of fine-tuning a pre-trained neural network model is one of the main contributions of this paper. In other words, we propose a method based on conditional GAN to regress the exact location of six body joints from input images concatenated with noisy and imprecise joint heat maps. 

\begin{figure}[t]
\begin{center}
   \includegraphics[width=1.2\linewidth]{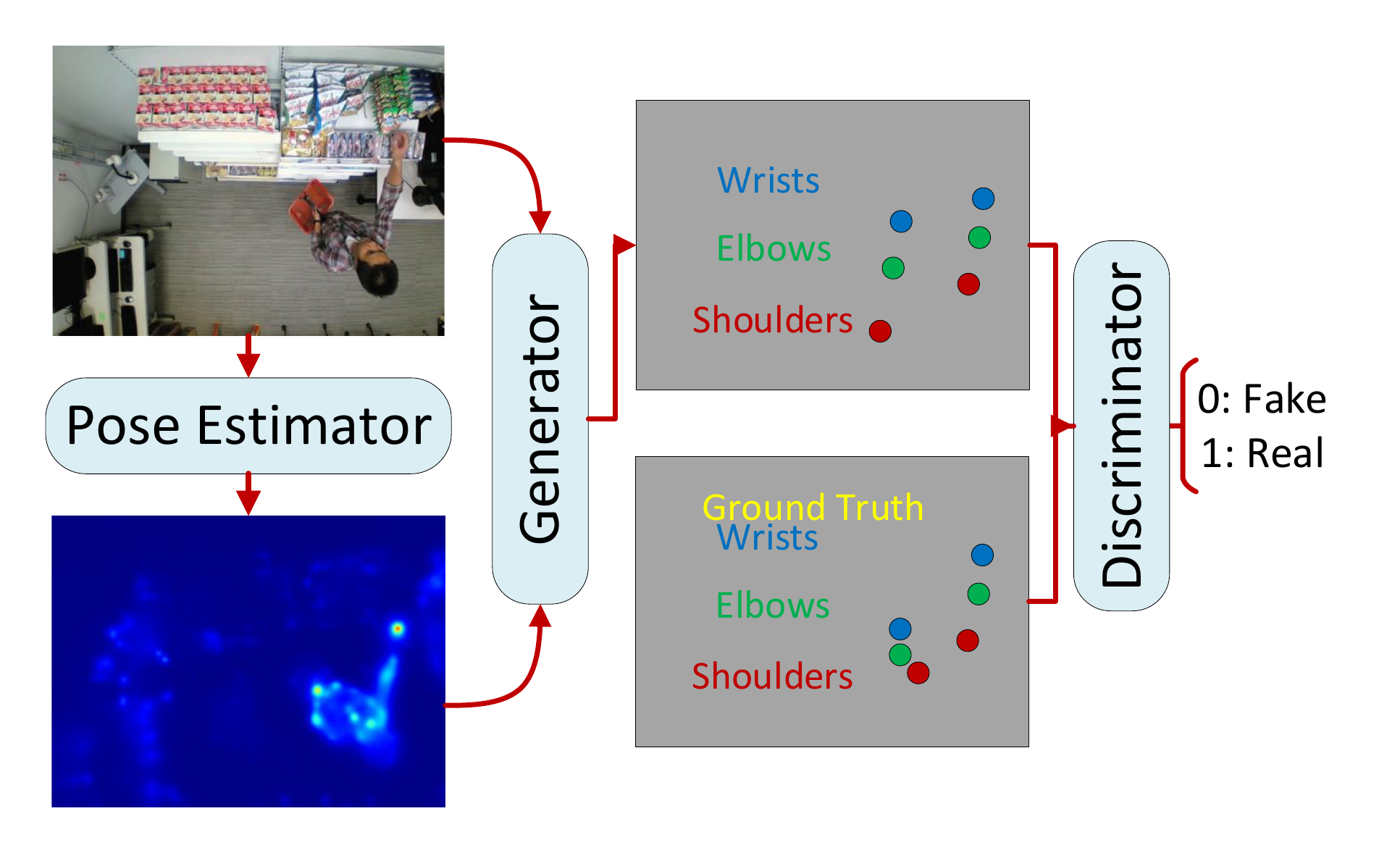}
\end{center}
   \caption{Overview of our GAN-based pose fine-tuning method.}
\label{fig:gan_arch}
\end{figure}

\begin{figure}[t]
\begin{center}
   \includegraphics[width=\linewidth]{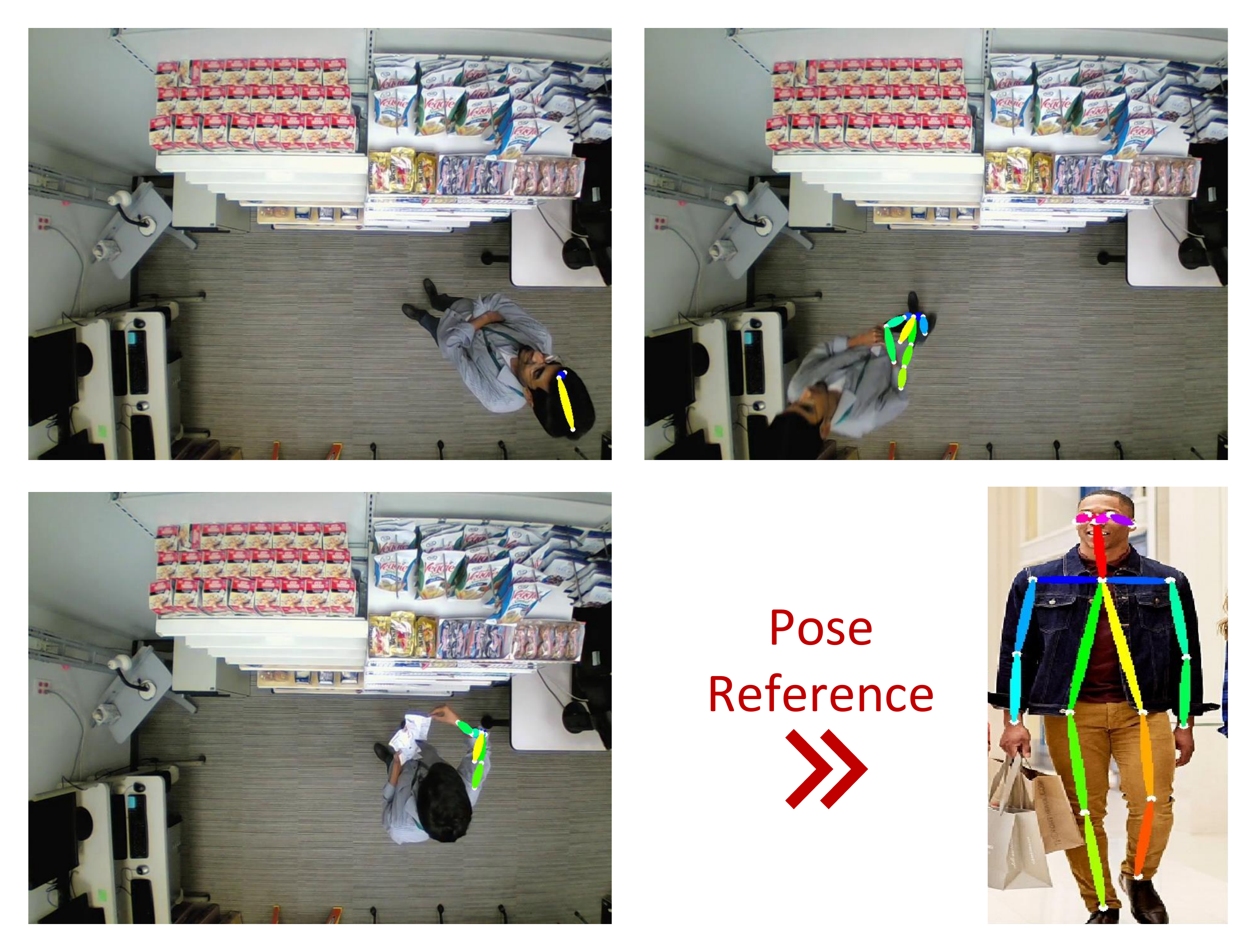}
\end{center}
   \caption{Examples of poor pose estimation results on MERL dataset.}
\label{fig:pose_example}
\end{figure}

As mentioned before, the camera viewing angle imposes a significant challenge to many computer vision tasks. In the shopping environment specifically, we often encounter top-view (or near top-view) recorded surveillance videos that further increase the difficulty of human activity understanding by imposing further occlusions of body joints and parts. This unique and challenging viewpoint has forced many deep learning researchers to avoid training on these type of input images. The pose estimator network that we incorporated in our task also is suffering from the same problem, even though having achieved state-of-the-art results on many other tasks that follow the routine camera view angle. Examples of the poor performance of the mentioned method in the current working dataset are illustrated in~\ref{fig:pose_example}. 

In order to improve the pose estimation accuracy and more accurately estimate joints-of-interest locations in our dataset, we propose the GAN structure shown in our approach overview~\ref{fig:overview}. As is shown, here the generator, which is a CNN with a similar architecture to VGG-11 followed by a multi-layer perceptron, is responsible for learning the conditional probability of the joint locations given the current extracted noisy heat map as well as the input frame (stacked). On the other hand, the discriminator is a Multi-Layer Perceptron (MLP) that is trained to distinguish between the real joint locations and the fake ones produced by the generator. This way we encourage the generator to produce more realistic joint locations and thus learn the conditional probability $ P (z | \theta)$, where $z$ is the stacked four channel input of heat map and original frame and $\theta$ is the output vector of six joint locations. In this structure, we follow the routine suggested in Wasserstein-GAN~\cite{wgan} to train our network in a stable manner. Since training WGAN with Gradient Penalty (GP) was shown to be more stable we follow this approach. Therefore, the overall optimisation problem for our WGAN with GP can be defined as follows:
\begin{eqnarray}
    L &=& E_{\Tilde{q}\sim p_g}[D(\Tilde{q})] - E_{q\sim p_r}[D(q)] + \nonumber\\ 
    &&\qquad\qquad\lambda E_{\Tilde{q}\sim p_{\Tilde{q}}} [(||\Delta_{\hat{q}}D(\hat{g}) - 1 ||)^2]
\end{eqnarray}

\begin{figure*}[t]
\begin{center}
   \includegraphics[width=\linewidth]{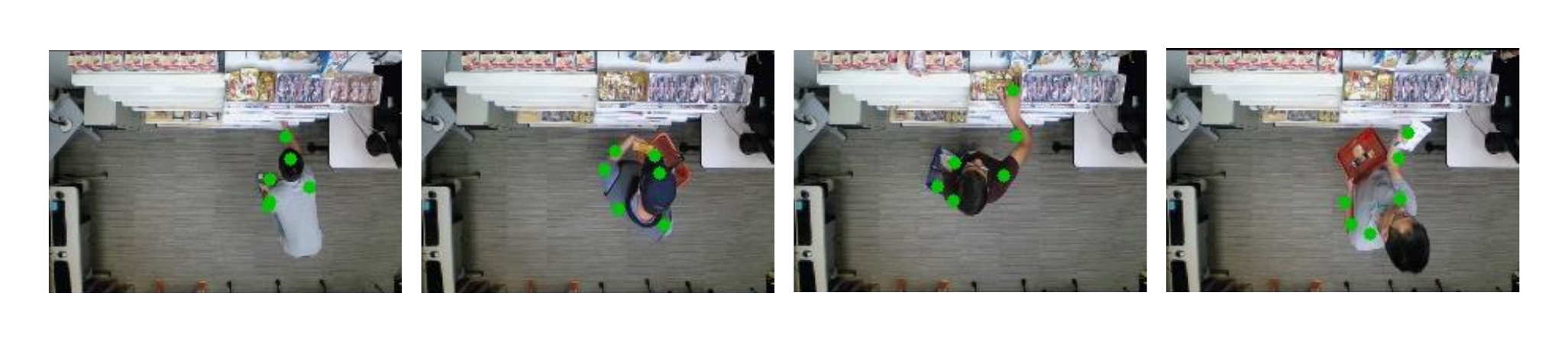}
\end{center}
   \caption{Joint location estimation results using GAN-based approach.}
\label{fig:gan_result}
\end{figure*}

Where $q \in R^{12*1}$ (1 is replaced with the number of frames accordingly), $p_g$ is implicitly defined in $\Tilde{q} = G(I)$, where $I$ is the input four channel image sampled from the dataset. (stack of heat map channel and the original frame).
In this framework, the trained generator is separated after being trained to use as a joint location regressor. In~\ref{fig:gan_result} we show some examples of how we infer high-quality joint locations given the imprecise and full of noise input heat maps.
Regarding the training, the network was trained for 200 epochs with learning rate of $10^(-4)$, number of critics set to five, $\beta_1 = 0.5$ and $\beta_2 = 0.99$. The generator was initialised using weights same as a weakly supervised pre-trained network on less than 0.5\% of the whole dataset frames, while the discriminator was initialised using Xavier random initialisation method.

\subsection{How to Incorporate Pose}
There have been many different methods proposed recently based on the latest success of pose estimation from RGB images to improve activity recognition and/or detection. The main flaw in most of them, however, stems from two roots, we believe. Firstly, treating the pose as only a combination of two-dimensional joint (or limb) locations is not a reliable clue for action recognition due to the natural differences between different joints and dismissing some crucial information such as the relative angles, direction and velocity of movement and relative position of them. Secondly, not all joints always play a similarly significant role in distinguishing the actions performed by the person. As in our case, there are only six joints that can be considered crucial in the current scenario happening in the shopping environment. Namely, both left and right shoulders, elbows and wrists.

Based on the above argument, we, in the first step, treat human pose as a probabilistic heat map of joint locations which we need to pay more attention to. We then propose that instead of regarding pose information as an independent source of information to define an action (which is impossible in some cases like our case even by using the perfect grand truth pose labels) we can add this source as a high-level understanding of the moving elements in the scene that enforces the neural network to extract features with more emphasis on these locations in the input frame, rather than general features extracted with huge amount of useless background data. This way we aid the network in capturing motion data without using any other specific motion representation, like optical flow which is too computationally expensive. Next, by adding the GAN fine-tuning step, we further remove the extra noise in the input heat maps and generate exact location maps of the six interest joints, which are fed as a substitute for the pose heat map channel to our pose stream network. It has been shown that stacking several video frames as input is an effective way to encourage motion capture by extracting features with a notion of movement in the scene. Our four channel input (three channels for RGB image and one for joint location map) is theoretically the same and encourages the CNN to learn the correlation between input channels.

In order to accomplish this task, we use the pose estimation algorithm proposed by Cao \etal~\cite{pose}. In their approach, the authors introduce a novel representation of body limbs as a normal vector of directions. In this way, the algorithm produces more accurate joint connections to generate a full body pose output. The network architecture that they propose to achieve this, is a multi-stage (optimised at six) CNN in which later stages are supposedly capable of detecting more global joint interactions rather than local. Therefore, each stage improves the pose estimation results of the previous one. The direct output of the proposed architecture is joint location heat map (probability maps) as well as limb direction heat maps, one per joint and two per limb. Then during post-processing, the heat maps are used to infer the human body pose of all the present people in the scene. 

Having visualised the final output of this approach, as is depicted in~\ref{fig:pose_example}, we observe that, most probably due to the lack of available training data on rare top-view images, the method performs poorly on our current dataset. Thus we propose the previously explained GAN-based approach to circumvent manual data annotation and fine-tuning this network. Additionally, our proposed method can potentially be used as an alternative in many similar fine-tuning tasks.

\subsection{From Recognition to Detection}
Intuitively speaking, our visual system as human-beings, is a nearly perfect action recogniser that can also detect the starting frames of actions in a video (frame sequence) effectively. We follow our intuition to provide our framework with the same capability. That is, we aim at creating a neural network that recognises single action having seen the whole sequence length effectively, then we combine its recognition knowledge with probabilistic reasoning to achieve start and end frame detection too. Therefore, we trained our network with shuffled trimmed sequences of actions instead of long untrimmed and sequential actions with a specific order. This way we remove the action-action sequence order supervision that clearly exists in the current working dataset as shown in~\ref{table:trans}. Therefore, we provide a more robust and reliable framework for real-world scenarios in which this sequential order might not exist.

\begin{table*}[]
\centering
\resizebox{0.6\linewidth}{!}{%
\begin{tabular}{l|c|c|c|c|c}
\textbf{Actions} & \textbf{Reach} & \textbf{Retract} & \textbf{Hand in} & \textbf{Inp. Product} & \textbf{Insp. Shelf} \\ \hline
\textbf{Reach} & 0.0 & 65.6 & 31.3 & 0.4 & 0.7 \\ \hline
\textbf{Retract} & 19.4 & 0.0 & 0.5 & 44.76 & 35.41 \\ \hline
\textbf{Hand in} & 0.084 & 88.3 & 0.0 & 0.2 & 0.6 \\ \hline
\textbf{Insp. Product} & 63.04 & 2.5 & 0.0 & 0.0 & 31.91 \\ \hline
\textbf{Insp. Shelf} & 99.1 & 0.0 & 0.2 & 0.7 & 0.0
\end{tabular}%
}
\caption{Transition matrix of actions in MERL dataset, using which reduces real-world applicability of every approach dramatically.}
\label{table:trans}
\end{table*}

In this approach, we use a fixed size sliding window that moves forward through the whole given input video with pre-defined stride (or step size measured in number frames). The sliding window size and stride values are optimally selected from a list of choices using a brute-force search.

\begin{figure}[t]
\begin{center}
   \includegraphics[width=\linewidth]{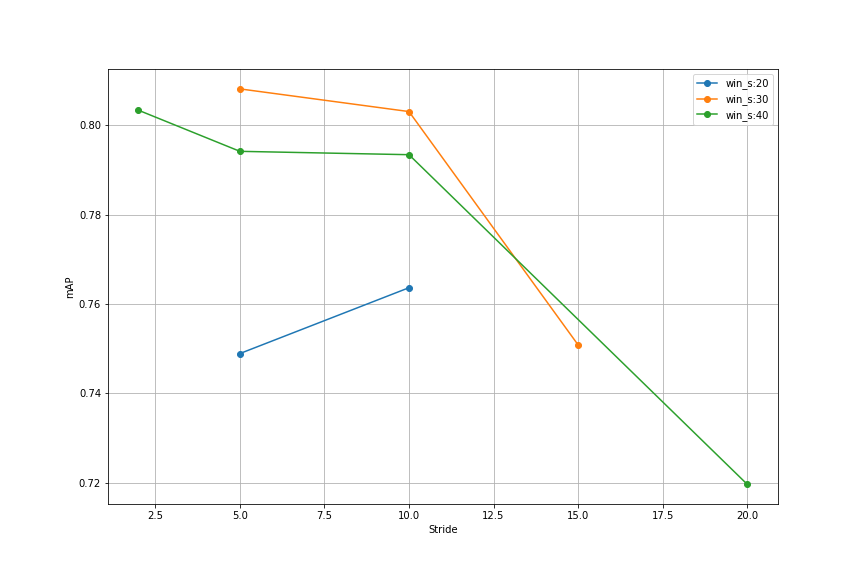}
\end{center}
   \caption{Optimisation of window size and stride value for best mAP.}
\label{fig:win_str}
\end{figure}

During inference, the label of each input frame is determined using the average probability of a total number of predictions as follows:
\begin{equation}
    Label = \text{argmax}(\frac{1}{N}\Sigma(P_i))
\end{equation}

\subsection{Object Motion} \label{obj_motion}
We argue that in many applications of activity understanding the movement of an object in the scene as a result of human-interactions is a crucial signal to determine the human activity, regardless of the type or category of it. Not only this applies to our environment of interest, \eg shopping, but it also holds correct for some other areas such as sports (holding a ball versus throwing it), and others too. Therefore, it's worth proposing a method that efficiently detects these object motion cues and fuses the attained information with other sources to decide on the action class.

Our object motion stream receives as input a fixed size bounding box map around the detected object and using three convolutional layers extracts features of the location, count and respective distance with interest points in the input frame. In order to detect the object, we follow the approach by Ren \etal~\cite{fasterrcnn}, so-called Faster R-CNN, and fine-tune the pre-trained network on our current dataset. Here we treat all the present objects as one unified class since only motion by itself is a strong cue for action recognition. For retail proposes, though, we plan to detect the category of the product too, as a further extension.

\subsection{Attention Modules}
In our proposed architecture we benefit from two different types of attention mechanisms. First, one is intended to increase the weights of extracted features in certain regions of input frames across the input sequence in a correlated way so that the attention region is dynamically moving with the movement of the human body, object or interest points specifically. The second one is to emphasise on more informative input frames in a given sequence of frames. Both of the modules are fully differentiable and learned during the network training. 

The first spatiotemporal attention follows the proposed approach in\cite{att}. According to their approach, the following module can be plugged in between convolutional layers of a CNN to modify the intensity of feature maps in salient regions across the input sequence:
\begin{equation}
    y_i = \frac{1}{C(x)}\sum_{\forall j} f(x_i, x_j)g(x_j)
\end{equation}

\begin{figure*}[t]
\begin{center}
   \includegraphics[width=0.8\linewidth]{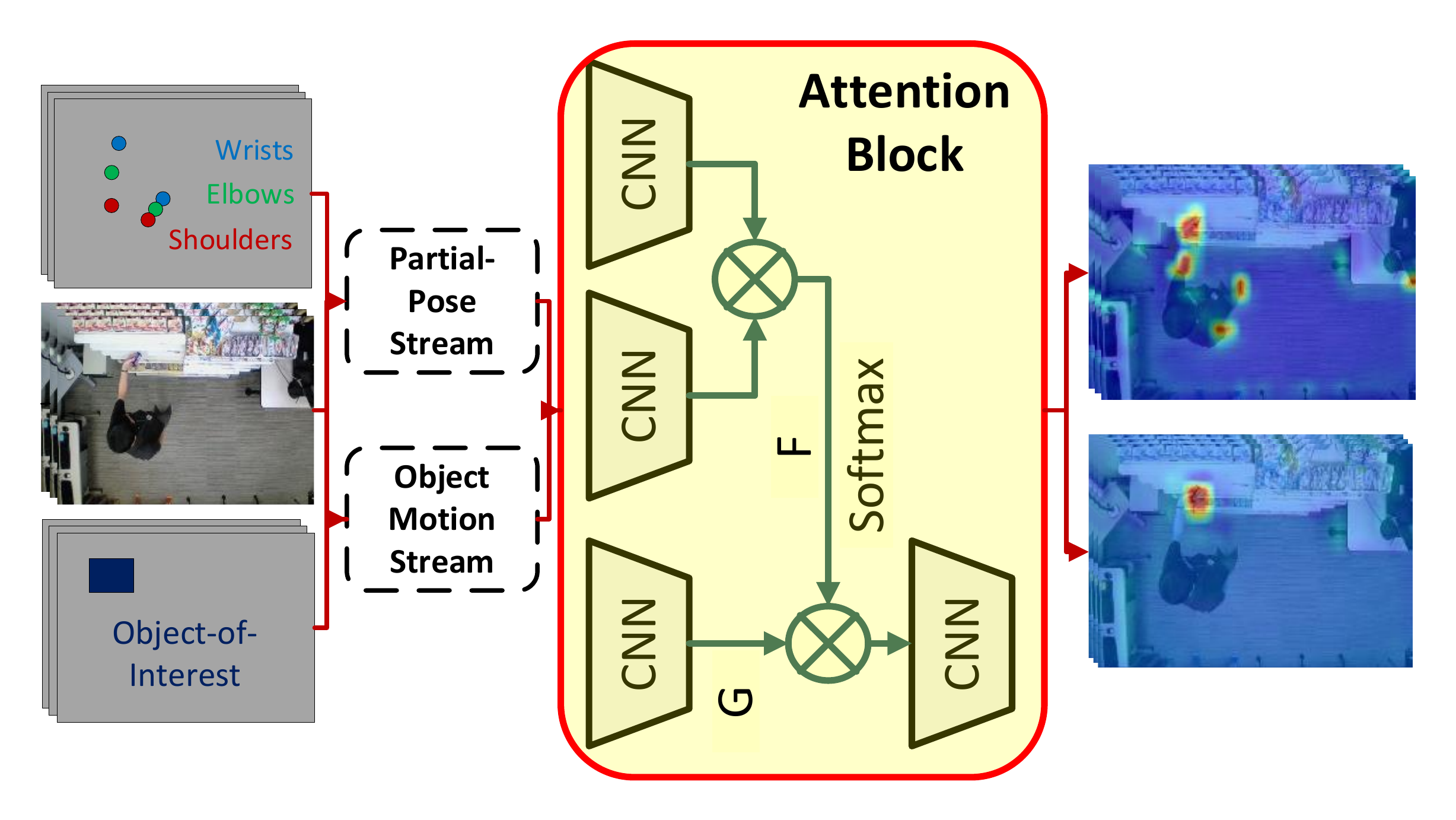}
\end{center}
   \caption{A Block of spatial-temporal attention and sample outputs demonstrating its effect on the input sequence of an action in both streams.}
\label{fig:att}
\end{figure*}

Where, as shown in the~\ref{fig:att} too, $f$ is an embedded Gaussian mapping of the input sequence, $i$ and $j$ are indices of the locations in frames across the whole sequence (space-time) and $g$ is a linear mapping function. 

Where $x_i$ is the input weight for a given position in space-time which is a function of all other possible positions $j$. The function $f$ that we use here is the Embedded Gaussian function:
\begin{equation}
    f(x_i, x_j) = e^{\theta(x_i)^T \phi(x_j)}
\end{equation}

We have empirically chosen the number of modules plugged into our network for best performance, which is two.
The second attention mechanism is applied to hidden state vectors of the LSTM. It basically associates a scalar weight with each input frame according to the importance learned by the network. In practice, the weights are applied to the hidden feature vector of the LSTM as follows:
\begin{equation}
    h_{out} = \sum_{i}w_ih_i
\end{equation}
In the equation above, $h_i$ is the hidden state of the LSTM at time step $i$ (corresponding to feature vector of frame $i$) which is multiplied by the weight value assigned by the temporal attention. In practice, this is performed by training a single fully-connected layer on hidden states of the LSTM.
These two modules, more importantly, the first one, have played a significant role in guiding how to improve our approach. As can be seen in the visualisations provided in~\ref{fig:att}, a higher weight is associated with the location of joints, specifically the six ones that build our partial pose model. We forced even more attention and consequently improved our results, by replacing the heat map of joint locations with precise maps generated by our generator network. In this way, we encouraged the concentration of the network resources with higher certainty by removing the noisy heat maps. 

\section{Experiments} \label{exp_ablation}

\begin{figure*}[t]
\begin{center}
   \includegraphics[width=\linewidth]{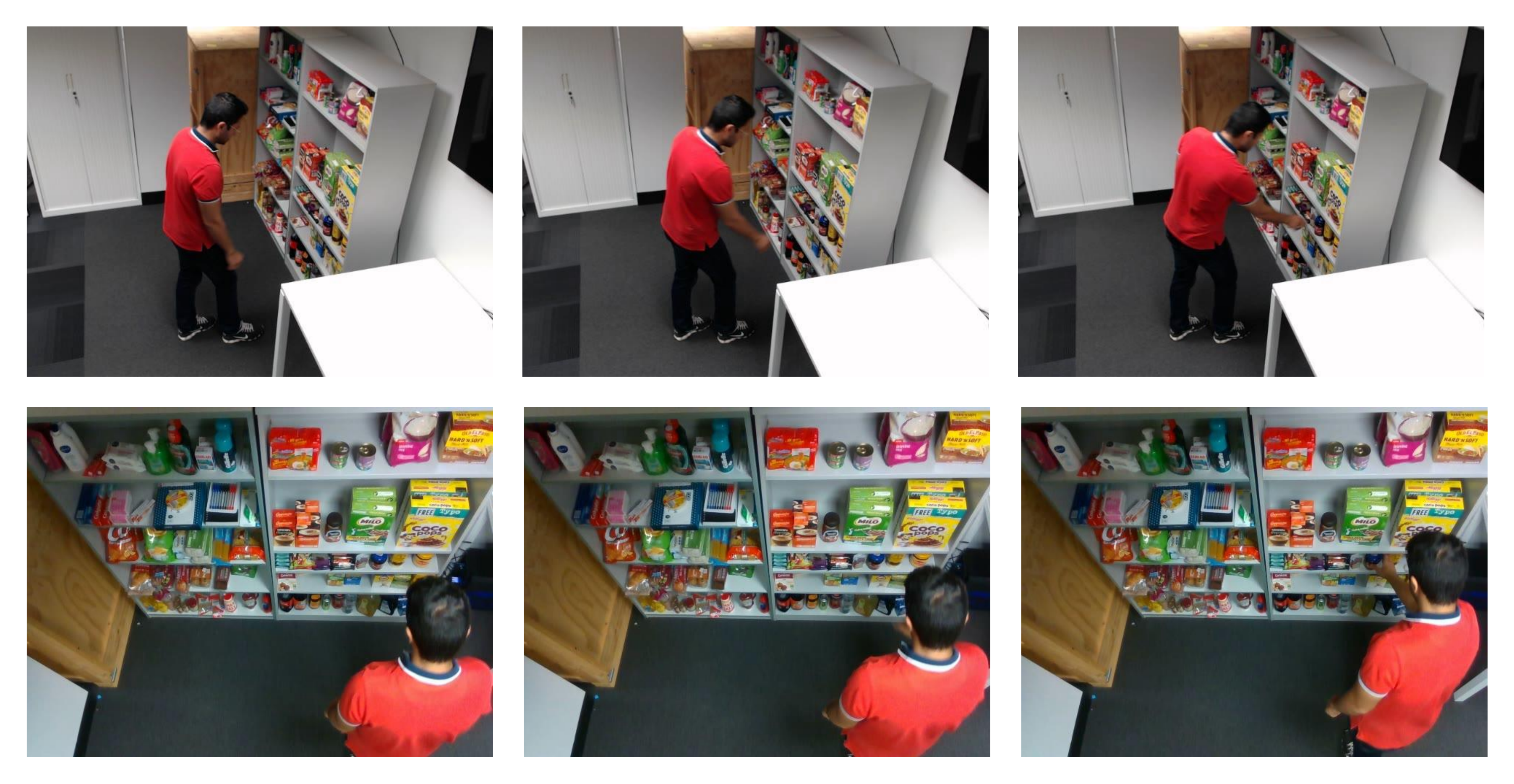}
\end{center}
   \caption{Sample frames of our dataset from two available camera views common in shopping environment.}
\label{fig:aiml_ex}
\end{figure*}

\begin{figure*}[t]
\begin{center}
   \includegraphics[width=0.7\linewidth]{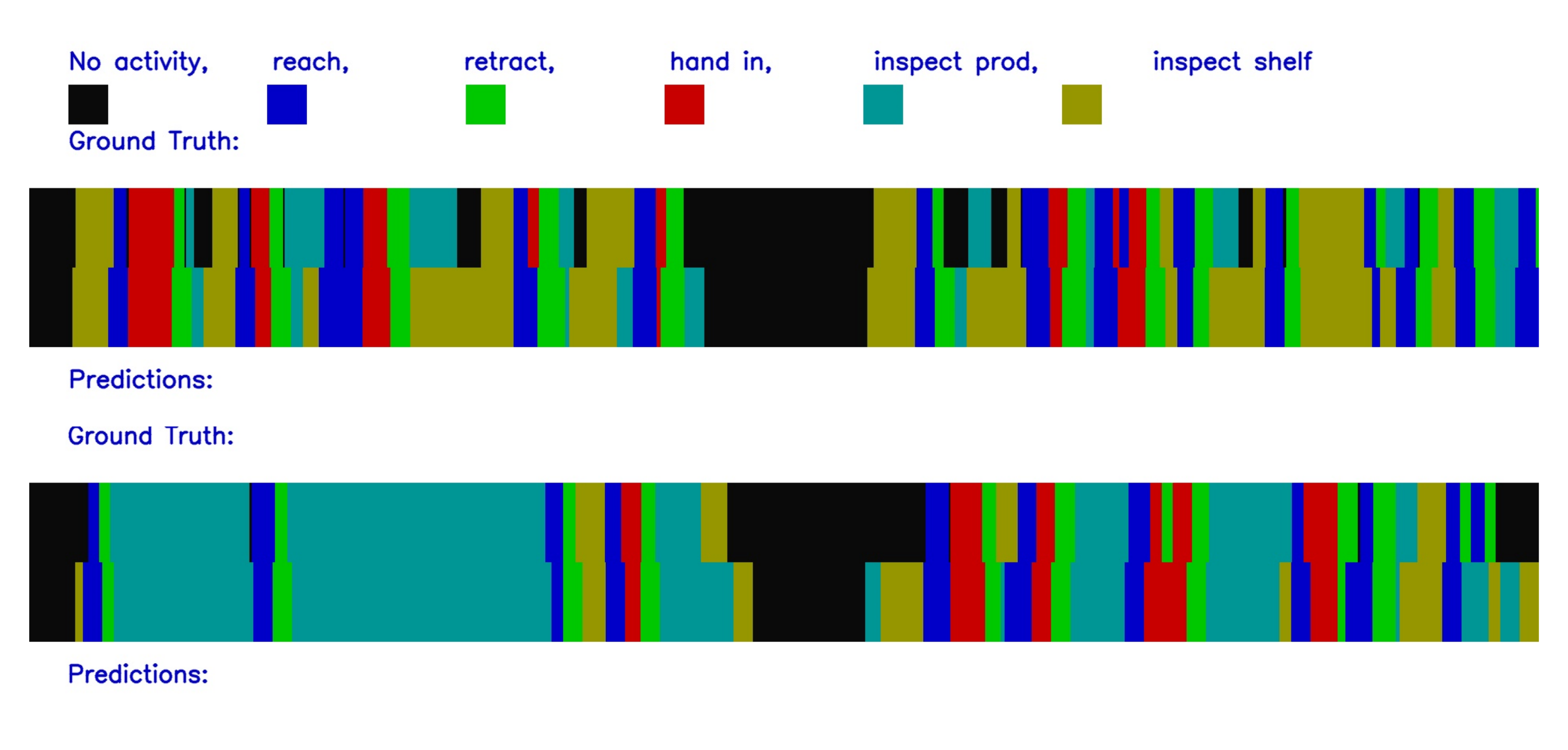}
\end{center}
   \caption{Visualisation of our detection results on MERL dataset.}
\label{fig:detection}
\end{figure*}

\subsection{Setup}

In order to fine-tune Faster-RCNN on our dataset almost 1000 uniformly sampled frames (from the total ~150,000 training frames) were manually annotated by a bounding box around the objects-of-interest. It should be noted that the object-of-interest comes from the fact that we’re not interested in all the objects in the scene; rather only the ones held at hand are the ones that define the interactions through their movements. Thus, in each frame, only if an object (\eg a product) is in close proximity of the hand is annotated.

Here we use the publicly available PyTorch implementation of the method. The fine-tuned network is then used along the rest of the network without further training to extract the prior knowledge regarding the object locations only.
In order to initialise the weights for the generator in GAN architecture, we manually annotated another 1000 uniformly sampled frames each with 6 joint-f-interest (\eg wrists, elbows and shoulders) locations. Then the generator is pre-trained in a supervised approach using the Mean Square Error loss function over the ground-truth joint locations. The discriminator, on the other hand, was initialised randomly using Xavier method. When the GAN training is converged, the generator is separately used along the rest of the network without further fine-tuning, as discussed before. 

The pose and object maps are binary maps generated in the following way. For joint locations, the map has values of $1$ in a circular area of constant radius (here 10 pixels) around each joint centre and $0$ elsewhere; similarly, the object map is $1$ in a rectangular area of fixed size (here 40 pixels) around the centre of each detected object and $0$ elsewhere.

The LSTM frame-wise attention module is implemented as a single Perceptron that provides a Softmax weight over each input frame that is then multiplied by the hidden state of the LSTM corresponding to that specific frame. This way the final classification is performed on a weighted average of all the hidden states rather than the last one only. This method provided us with considerable accuracy improvements over simple last hidden layer classification.

Since the inputs, sequences of video frames containing a single activity, have various lengths with high variance (ranging from short activities of 1-sec length to long activities of above 10 seconds) there are several strategies to unify the length of sequences so that they can be fed into LSTM. One strategy is to zero pad the shorter inputs to the length of the longest one. This is not a good idea, in our case, since the high variance of input size means sparsity of real data among the padded zeros in some cases that makes learning hard for the LSTM. The other strategy is to deterministically, or even randomly, sub-sample all the sequences to the same defined size. This way finding the sweat equilibrium point between losing the important frames and reserving the required number of frames is not trivial. Therefore, instead, we use the original input sequences without any change and only specify a hard limit (here 40 frames at 15 frames per second rate), where the longer sequences are sub-sampled to 50\% until satisfying this limit.

Having varying input size, however, renders batching the input impossible. Thus, we use a commonly used technique in training neural networks called Gradient Accumulation. That is, during the training, we accumulate the gradients calculated on each input sequence for the defined mini-batch size (here 12) before back-propagation. This is, essentially, a simulation of mini-batch training that effectively achieves the desired training outcome.
The whole recognition network is trained using gradient descent to minimise the Cross-Entropy loss over the Softmax class probability outputs, given in the following formula:

\begin{equation}
    Loss = -\sum_{i}^{C} y_i log(p_i) 
\end{equation}{}

Where $y_i$ is the ground-truth activity label and $p_i$ is the output of the network. Cross-Entropy has been proven useful in practice for classification tasks by nonlinear penalisation of far-off probability scores.

The optimiser used here is Adam that has been proven to provide higher performance in large scale models, like ours, in terms of convergence rate. This is achieved by dynamic (\eg adaptive as the authors call) adjustments of the learning rate for each weight based on the first and second order moments of the gradient thus far. The hyper-parameters chosen here are the default values suggested by the authors as $\alpha = 0.001$, $\beta_1 = 0.9$, $\beta_2 = 0.999$ and $\epsilon = 10^{-8}$.
Training the spatiotemporal attention modules in intermediate layers of the two streams is a tricky task. These modules were assigned a contribution weight gamma (consistent all around the network), initially set to $0$. After convergence of the main network, we start the training again for a few epochs (~7 to 10) with an initial $\gamma = 0.1$ which then gradually is increased to $1$. 

Finally, given the fact that each stream has a similar architecture to VGG-11 network, we initialise both the streams using transfer learning from the networks trained on COCO classification tasks. This way we are able to improve the training length considerably; that is the whole network training converges in around 15~17 epochs.

The last step, then, is to optimise the sliding window and stride values for detection. This is done using the Brute Force search in two given sets of values. For each pair of window-stride size value, the overall performance is evaluated, as discussed in the evaluation section, and the best performing values are selected. The sliding windows sizes range from $5$ to $40$ frames while the stride ranges from 1 to the size of the sliding window.

The whole training and inference are implemented in Python using PyTorch deep learning package.

\subsection{Evaluation}
We evaluate our proposed method and some variants on the two introduced datasets for recognition accuracy and detection precision. The former assumes the given video contains a single action while the latter receives a long video (and unlimited sequence of frames) of multiple actions as input and the objective is to temporally localise and classify the actions using a single forward feed. The first metric we use for evaluating detection is the common $F_1\text{score}$. This metric provides a rather fair quality of the detection by evaluating both the precision and recall scores and consequently penalising both for false positives and false negatives. It can be defined as follows:
\begin{equation}
    F_1\text{score} = \frac{2 * \text{Precision} * \text{Recall}}{\text{Precision} + \text{Recall}}.
\end{equation}

In addition, we report our Intersection over Union (IoU) minimum threshold that differentiates between a true positive and a false positive instance in detection. We argue that the higher this value is selected the less over-segmentation scores are assigned to the detection performance. Therefore, reporting high performance for IoU values under 0.5 is essentially meaningless due to the high localisation error. 

\subsection{MERL Datset}

MERL dataset is collected to address the lack of appropriate datasets in shopping environments. It is labelled by six activities, namely: 'reach the shelf', 'retract from the shelf', 'hand in the shelf', 'inspect the product', 'inspect the shelf' and the background (or no action) class. It's recorded using a roof-mounted camera to simulate the real shopping store environment, albeit not specific. There are $42$ participants in this dataset as shoppers. One particular drawback of this dataset is that the participants are instructed to perform a sequence of tasks that one can easily exploit as well-behaved transition probability matrix to achieve good results on this dataset at the cost of over-simplification. An example of this drawback is shown in table\ref{table:trans}. This provides the network with a good prior knowledge which can be incorporated during inference to increase the accuracy of recognition. Nevertheless, one major advantage of our approach is that we achieve state-of-the-art results without relying on this strong prior knowledge.  This is very intuitive as in real-world solutions, such strong priors are hard to come by and unrealistic, thus cannot be deployed. Therefore, in our released dataset (will be discussed in the subsequent section), we removed these biases by ensuring the participants are not following a simple sequence of activities.

\subsection{AIML Shopping Activity Dataset}
In this dataset, we tried to alleviate the problems that we faced while working on MERL dataset. Some of the main issues include some longer than usual actions, sparsity and low frequency of actions and some instances of imprecise labelling. Additionally, we provide another view of the scene which is also common in shopping environments as well as depth information from top-view. Examples of action sequences from both views are provided in figure~\ref{fig:aiml_ex}. 
In this dataset, which has been collected in an encouraging way for the participants to act more naturally, we propose more challenges to the detection task. This challenge can, for example, be observed in the number of actions that on average happen per minute; statistically speaking, in MERL we deal with $21$ actions per minute on average while in our dataset the average is around $35$ per minute. This higher frequency and shorter length of each action, which we believe is more realistic, imposes more challenges, too. 
In overall, this dataset contains videos of $20$ people performing normal shopping activities consisting of the classes defined in MERL dataset. The shoppers either follow their own instinct to perform activities habitually once. In at least two other cases, each person is provided with a shopping scenario, like shopping for a small party or weekend breakfast. The simulated shopping environment consists of two shelves filled with more than $150$ different products. This provides more generalisation capability to our object detection as well as a more realistic to encourage real-life activities.
The results of applying our method to this dataset are shown on~\ref{table:own_res}

\begin{table}[]
\centering
\begin{tabular}{l|c}
\textbf{Measure} & \textbf{Value} \\ \hline
\textbf{Frame-wise acc.} & 58.88 \\
\textbf{F1 @ 50} & 65.27 \\
\textbf{Overall recognition acc.} & 68.76
\end{tabular}
\caption{Our results on our newly intrduced dataset. The new imposed challenges require improvements in the approach.}
\label{table:own_res}
\end{table}

\begin{table}[t]
\resizebox{0.99\linewidth}{!}{%
\begin{tabular}{lcc}
\textbf{Method} & \textbf{F1\{IoU=0.5\}} & \textbf{Frame-wise Accuracy} \\ \hline
\textbf{Singh \etal~\cite{merl}} & 65.4 & 76.3 \\ \hline
\textbf{Lea \etal~\cite{lea_merl}} & 72.9 & \textbf{79.0} \\ \hline
\textbf{Bai \etal~\cite{merl2018}} & 74.8 & 77.1 \\ \hline\hline
\textbf{Ours} & \textbf{76.02} & 74.7 \\
\end{tabular}}
\caption{Detection results on MERL dataset.}\label{table:f1}
\end{table}

\subsection{Results}
In this section, we provide the state-of-the-art results achieved on MERL dataset as well as the first benchmarking results on Our dataset for the benefit of the research community. Unlike previous approaches reporting on MERL dataset, we both provide recognition as well as detection results. The former assumes the input videos contain a single action only while the latter receives a long video of multiple actions as input, so-called untrimmed video. Table~\ref{table:f1} shows our recognition results on MERL and table compares our detection results with previously reported ones. Since the recognition results on this dataset have never been reported before by any other method (to the best of our knowledge) this will open up a new challenge for the future research so that other later algorithms can compare their recognition results. As we have seen and is reflected in our results too, for a reasonably good detection method, the precision of detection should be close to the accuracy of the recognition. 

As can be seen compared to other approaches, we significantly have higher IoU (average 0.67 compared to 0.5 as the maximum reported) over previous approaches while achieving better detection results, too. However, the significance of our method is above and beyond based on the following argument:
All other methods achieve their results by designing an a-casual approach. That is a system that decides not only using the past seen frames but also by benefiting from the future unseen ones, too. This, we believe and argue, that is totally unfair and unrealistic, especially in the shopping environment. An a-casual system is no practical advantage and is not deployable in practice.
Furthermore, by accessing the future unseen frames, these methods can easily benefit from another source of valuable information that is in practice inaccessible. That is the transition matrix provided in table~ that provides prior knowledge of current activity given the previous ones.
Our method, on the other side, is independent of both of these unfair sources of information is certainly deployable in practice.

The frame-wise accuracy of our method is slightly lower. However, we argue that this metric in action detection is not a very suitable measure of performance of for two main reasons. Firstly, most of the accuracy we lose is due to a lead or lag between our network's prediction start or end frame and the grand-truth, which is to some extent inevitable. This can be seen in~\ref{fig:detection}. Secondly, having visualised the instances of wrong predictions, we have noticed some inaccurate labelling in the MERL dataset which influences our results. 
Also, the optimisation of the sliding window size and stride value graph is depicted in the~\ref{fig:win_str}. This has a straight relationship with the average length of action sequences that were provided before.

The influence of both attention modules is also shown in example images. In order to illustrate the attention weights we follow \etal that apply the following formula to extract per pixel weights associated with the spatiotemporal attention module:
\begin{equation}
    W_{i, j} = \sum_{\forall m} ||W_{i, j}||_m 
\end{equation}
Where $W_{i, j}$ is the associated weight to a specific pixel and $ m $ is that location across all the available channels.

\subsection{Ablation Study}
Here we provide more insights into our proposed approach that empirically proves the contribution of each component; this also demonstrates the necessity of the presence of these components in our framework to enable incorporation of all the available sources of information for more accurate inference. This is all summarised in table~\ref{table:ablation} which is discussed in detail in this section.

\begin{table*}[]
\centering
\begin{tabular}{l|ccccccc}
\textbf{Method/ Class} & \textbf{No Act.} & \textbf{Reach} & \textbf{Retract} & \textbf{Hand in} & \textbf{Product} & \textbf{Shelf} & \textbf{Overall}\\ \hline
\textbf{Raw Frame} & 87.50 & 87.84 & 91.40 & 87.50 & 68.12 & 81.12 & \textbf{83.91} \\ \hline
\textbf{Pose Stream} & 85.23 & 93.30 & 91.67 & 83.75 & 71.10 & 85.71 & \textbf{85.11} \\ \hline
\textbf{Object Map} & 81.81 & 78.41 & 45.31 & 42.50 & 35.51 & 86.22 & \textbf{61.63} \\ \hline
\textbf{Bi-Stream} & 87.50 & 94.79 & 92.45 & 81.88 & 76.09 & 89.80 & \textbf{87.08} \\ \hline
\textbf{Bi-Stream + Att.} & 87.50 & 95.04 & 92.97 & 85.00 & 78.99 & 92.35 & \textbf{88.64}
\end{tabular}
\label{table:ablation}
\caption{Ablation study summary. This table empirically proves the significance of each component and effectiveness of integration method.}
\end{table*}

The first row shows the use of raw input frames as the input to a single stream of the network. Since the network is symmetric this can be any of the two streams. The LSTM is untouched as is the attention mechanisms and the same training procedure as before is followed. All in all, this row provides the best results that our proposed method can achieve when the only input is the raw frame. As can be seen, the values are relatively high which shows the effectiveness of the overall method including the attention mechanisms and avoiding zero-padding the inputs. However, this is still almost $55\%$ less than the full capacity of the method on average.

The second row follows the same scenario as the previous one with the difference that joint location maps are stacked along with the input frames. Again, the same training procedure is followed. This clearly shows the significance of joint location (\eg body pose) in activity recognition by improving the accuracy by around $2\%$. 

The third row is an empirical proof on the vitality of our proposed object motion stream in distinguishing human activities. The input to the network is only object maps generated as discussed in section~\ref{obj_motion}. As is demonstrated, object motion is a very strong clue that our method extracts efficiently.

Comparing the last two rows, finally, shows the significance of attention mechanism in activity recognition. As can be inferred, nearly $2\%$ of the overall accuracy can be related to the way attention modules are incorporated in our approach. Needless to say, as the title of this paper suggests, the attention is further encouraged in specific locations in the input frames. This is achieved using the special way that we integrated human pose and object motion.  

\section{Further Attention} \label{RL_att}

Since the first integration of deep learning into RL in 2013~\cite{rl_mnih, rl_mnih2} leading the impressive success in playing Atari games, many researchers have been motivated to apply RL to more challenging tasks and further exploit the capabilities of this method. The tasks range from navigation~\cite{rl_navigation}, activity recognition~\cite{rl_activity}, image captioning~\cite{rl_captioning} to dialogue~\cite{rl_dialogue} systems or even object detection and tracking~\cite{}; there is essentially no limit.

Before applying RL to any task there are requirements to be satisfied. The main one is the problem being defined as a Markov Decision Process (MDP). A MDP is a tuple of $(S, A, P_a, \gamma, R_a)$; that is defining the set of states $(S)$, actions $(A)$, state transition state transition probability $P_a$, discount factor $\gamma$ and reward function $R_a$. The problem definition should also satisfy the Markov property in the sense that all the necessary information for decision making is available in each state. This makes the future independent of the past given each present state. Then the RL problem is about finding the optimal policy providing action distribution in each state.

In order to fit the attention mechanism problem into RL, we divide it into two different sub-problems. First, an MDP to decide on the salient region locations in each single input frame for the whole sequence of an activity (note that the activity is the person’s activity in the video while the action is the decision by the RL agent). Second, a weight adjustment mechanism that efficiently increases the intensity of feature maps using a pre-defined distribution (\eg simple Gaussian for the moment). This is certainly the theoretically safest problem definition that has enough chances of success. Therefore, later on, each sub-problem might be made more complicated with higher capabilities.

\begin{figure}[t]
\begin{center}
   \includegraphics[width=\linewidth]{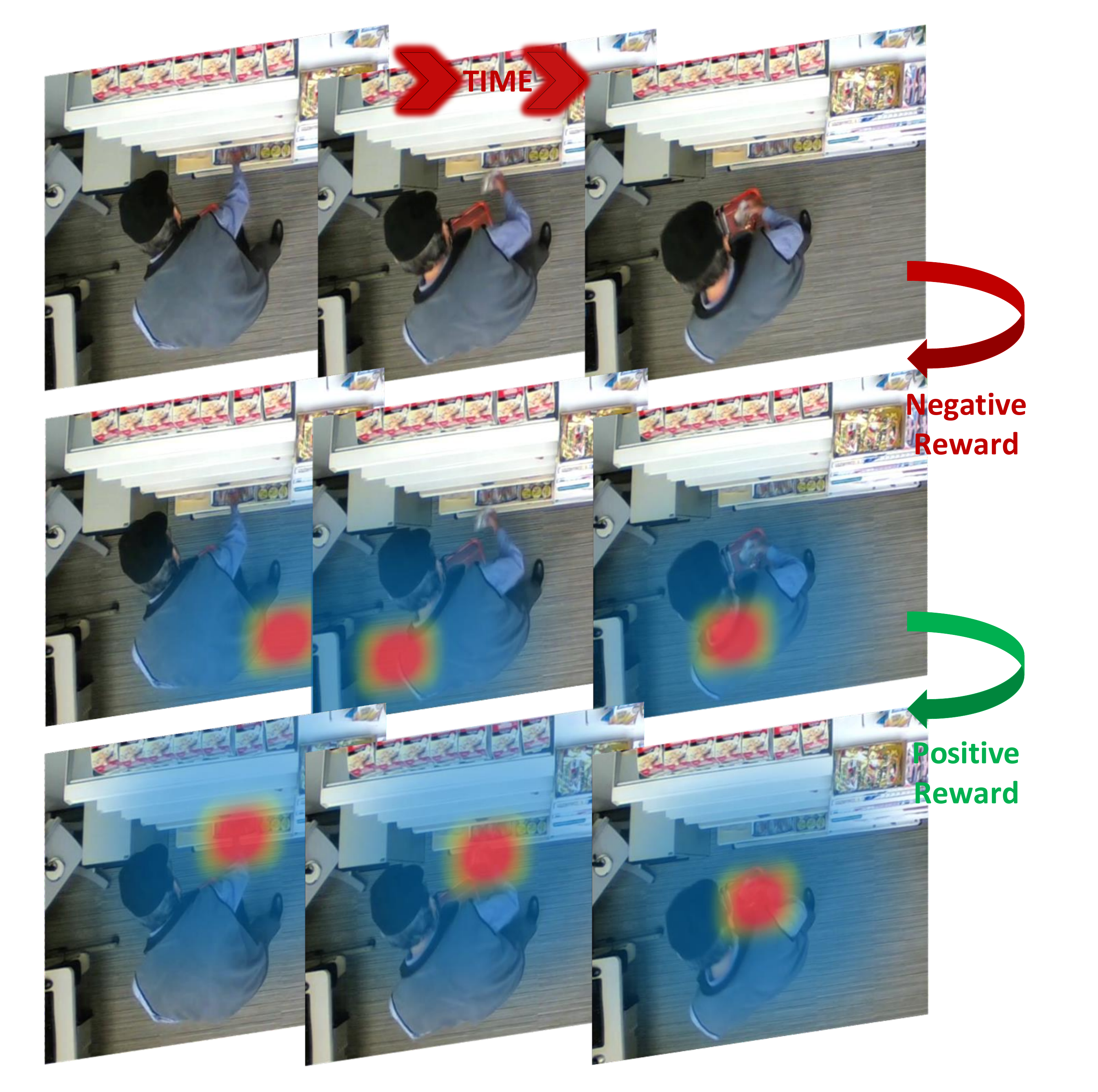}
\end{center}
   \caption{Visual introduction of our proposed RL-based attention mechanism.}
\label{fig:rl_att}
\end{figure}

The overall designed approach is illustrated in figure~\ref{fig:rl_att}. The state space consists of the whole set of input frames in the working dataset. An episode of the MDP is then an activity sequence consisting of several frames. Actions are trivially defined as the index of the salient grid in the grid world division of the input frame, initially. There are two possible choices for the neural network that plays the role of policy function approximator. First, a small CNN that receives a single frame and attends to various regions in different frames. Second, an RNN that selects salient regions that are correlated during an episode (sequence). In the latter, the state space should include the hidden states of the RNN, too, to be considered fully observable MDP.

Under any of the above scenarios, the reward function is a fraction of the change in the loss function (\eg Cross Entropy) with respect to the case where attention is not involved (pre-trained classifier network). The whole network could be trained using Policy Gradients~\cite{policy_g} where the back-propagation is done after rolling out some episodes (mini-batch).

\section{Conclusion}
We proposed a framework for activity recognition and detection in a fine-grained manner in shopping environments. The challenge in fine-grained detection is the length of each action being very short, intra-class variance being high and the inter-class variation being low; these challenges impose intrinsic difficulties to the task. Additionally, the extrinsic challenge of rare and unique camera viewing angle renders the task more demanding. we proposed a weakly supervised method using GAN to fine-tune pose estimation results when there is a difference between the camera viewing angle during training and inference. We provided extensive experimental results to prove the applicability of our method to real-world cases, specifically in shopping environments with all their unique characteristics and challenges. Finally, we release a new dataset for the challenging task of customer activity and behaviour understanding with this paper to the research community hoping to encourage further research that is currently limited to the companies with access to preparatory data.

\section{Future Works}

Despite having accomplished state-of-the-art results in activity detection in a shopping environment, we observe many potential extension possibilities and the prospect of improvement. However, due to the intensive time-limit, we leave them as suggestions for future work presented in this section.

Firstly, we believe that the human body pose has the potential to accurately define and distinguish actions, even as an independent stand-alone feature. However, the endeavour dedicated to this during our project proved it very challenging. The most significant drawback of current pose estimation approaches is ignoring temporal dependency between frames. We believe that combining pose estimation with the optical flow is one way to use the joint movement patterns as prior knowledge for predicting the location of joints in the next time step, where direct estimation fails (\eg occlusion).

Secondly, even by having an (almost) continuous flow of joint locations through time (\eg in a sequence), our initial experiments empirically showed that simple normalised location coordinates are not informative-enough signals for defining an action. Therefore, considering the similarity of body pose to a graph structure, we suggest using graph neural networks would improve the results.

Next, as a result of the previous extensions, using an auxiliary object detection module to detect the objects held at hand would be redundant. That is because having the exact wrist location available is enough to check the presence of the object at hand.

Additionally, the power of GAN in the last few years has become clear to the research community. We consider our proposed approach for joint location estimation from heat maps has high potentials to become a general approach for fine-tuning neural networks in an unsupervised manner for similar tasks (or many others by modification). Here, fine-tuning means improving the final results without further training the original network and thus in the absence of expensive annotations. 

Furthermore, with the recent surge of success in deep reinforcement learning methods, we suggest defining the attention mechanism as a salient region localisation task for an intelligent agent could potentially reduce the computational complexity of its training as well as inference. Besides, as can be seen in attention visualisations, using RL may lead to more focused attention and thus improve the results, too.

Finally, as discussed during the introduction, activity understanding is only one major building block of customer behaviour analysis. We suggest that further attempts on facial expression recognition and/ or eye gaze estimation provide complementary information for deep behaviour analysis of customers. We hope that gaining these insights, not only can help instantaneous anomaly detection but will also provide retailers with answers to questions such as: "What exactly attracts the customers in terms of price, packaging, shelf arrangement, staff scheduling etc.?"

\clearpage
{\small
\bibliographystyle{ieee}

\bibliography{main}
}
\end{document}